\documentclass[conference]{IEEEtran}
\usepackage{multirow}
\usepackage{array} 
\usepackage{graphics}
\usepackage{cite}
\usepackage{subfig}
\usepackage[pdftex]{graphicx}
\usepackage[absolute]{textpos}

\TPMargin{5pt}

\newcommand{\copyrightstatement}{
    \begin{textblock}{0.84}(0.08,0.93)    
         \noindent
         \footnotesize
         \copyright~2022 IEEE. Personal use of this material is permitted. Permission from IEEE must be obtained for all other uses, in any current or future media, including reprinting/republishing this material for advertising or promotional purposes, creating new collective works, for resale or redistribution to servers or lists, or reuse of any copyrighted component of this work in other works.
    \end{textblock}
}

\begin{document}

\copyrightstatement

\title{A reconfigurable integrated electronic tongue and its use in accelerated analysis of juices and wines}

\author{\IEEEauthorblockN{Gianmarco Gabrieli, Michal Muszynski, Patrick W. Ruch}
 \IEEEauthorblockA{IBM Research Europe, S{\"a}umerstrasse 4, 8803 R{\"u}schlikon, Switzerland}
}

\maketitle

\begin{abstract}
Potentiometric  electronic tongues (ETs) leveraging trends in miniaturization and internet of things (IoT) bear promise for facile mobile chemical analysis of complex multi-component liquids, such as beverages. In this work, hand-crafted feature extraction from the transient potentiometric response of an array of low-selective miniaturized polymeric sensors is combined with a data pipeline for deployment of trained machine learning models on a cloud back-end or edge device. The sensor array demonstrated sensitivity to different organic acids and exhibited interesting performance for the fingerprinting of fruit juices and wines, including differentiation of samples through supervised learning based on sensory descriptors and prediction of consumer acceptability of aged juice samples. Product authentication, quality control and support of sensory evaluation are some of the applications that are expected to benefit from integrated electronic tongues that facilitate the characterization of complex properties of multi-component liquids.
\end{abstract}


\IEEEpeerreviewmaketitle

\section{Introduction}

The design of cost-effective and rapid screening sensing systems is key to deliver alternative tools for chemical analysis, and electronic tongues have demonstrated to be potentially disruptive for various applications \cite{IEEEexample:podrazka2018electronic}. In particular, ETs have proven useful to identify food products, quantify their major constituents and possibly predict taste attributes \cite{IEEEexample:titova2020electronic}. In this context, potentiometric ETs offer clear advantages, such as facile integration and fast response, that make them suitable candidates in the food industry \cite{IEEEexample:ciosek2011potentiometric}. Hereinafter, an existing technology platform \cite{IEEEexample:ruch2019portable} including a microcontroller-based data acquisition, smartphone interface and cloud computing back-end is combined with a miniaturized array of 16 electrodeposited conductive polymers and a specific training methodology \cite{IEEEexample:gabrieli2021} to further extend its capabilities. The present contribution demonstrates how  different machine learning models can be deployed to reconfigure the same device to analyze fruit juices and bottled wines, and how discrimination of liquids is achieved based on their chemical and sensory attributes. 

\section{Methodology}

\subsection{Sensor array and liquid samples} 
\begin{figure}[h]
\centering
\includegraphics[width=2.6in]{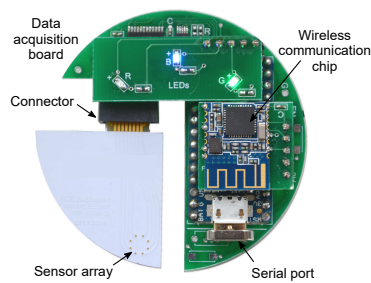}
\caption{Portable electronic tongue with integrated sensor array.}
\label{fig_device}
\end{figure}

A previously described~\cite{IEEEexample:gabrieli2021} potentiometric electronic tongue comprising 16 conductive polymeric sensors electropolymerized on a common substrate was used in all experiments (Fig.~\ref{fig_device}). The array was used to test acetic, citric and lactic acid at five concentrations between $10^{-5}-10^{-1}$ M. The pH of the solutions was adjusted with NaOH to maintain values between 4.2--4.5. A $10^{-3}$ M solution of the respective acid was used as reference solution in each test. The same array was then used to test nine fruit juices of different flavours (4$\times$ orange, orange-passion fruit, pear, peach, apricot and multivitamin), whereby one of the orange juice samples was used as reference liquid for testing. Finally, the sensor array was used to test eleven bottled Italian red wines, using  one of the wines (Palazzo della Torre, P.d.T.) as reference liquid.

\subsection{Measurement and data processing}
A total of 75 discrete hand-crafted features~\cite{IEEEexample:gabrieli2021} were extracted from 15 unique differential voltages time-series during transition of the sensor from the respective reference liquid to a test liquid. Tests were repeated in randomized order using an automated sampling system~\cite{IEEEexample:gabrieli2021}. One orange juice was tested after storage at 40$^{\circ}$C for 10, 20, 40 or 50 days and compared to the same juice stored at room temperature or in a refrigerator, whereby the latter was also used as reference liquid during testing. All liquids were tested at room temperature. A group of 12 panelists was asked to taste the same juices and express their willingness to purchase each of them (\emph{"purchase criterion"}). A sample was classified as \emph{accepted} when at least half of the panel members replied in favor of the purchase. All classification problems were tackled with three different machine learning algorithms: Linear Discriminant Analysis (LDA), K-Nearest Neighbors (KNN with $k=3$) and an ensemble of 50 bagged decision trees.

\section{Results and discussion}

\subsection{Cross-sensitivity to organic acids}

\begin{figure}[h]
\centering
\includegraphics[width=3.3 in]{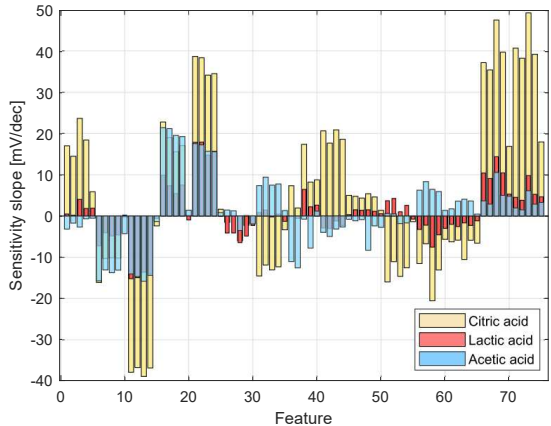}
\caption{Feature sensitivity to organic acids}
\label{fig_sense}
\end{figure}

Previous work had demonstrated sensitivity of the electronic tongue to metal ions~\cite{IEEEexample:gabrieli2021}. In the present contribution, each of the 75 features also exhibited variation with changing organic acid concentration, expressed as the slope of the feature magnitude against the logarithm of the acid concentration (\emph{sensitivity slope}, Fig.~\ref{fig_sense}).  Each group of five features is extracted from the same voltage channel comprising a specific pair of polymeric sensors. The feature response is clearly differentiated across the three acids, demonstrating cross-sensitivity of the sensor array. This result suggests that the E-tongue interacts with organic constituents relevant for many beverage classes, including fruit juices and red wines.

\subsection{Unsupervised clustering analysis}

\begin{figure*}[h]
\centering
\subfloat[Fruit Juices]{\includegraphics[width=3.4in]{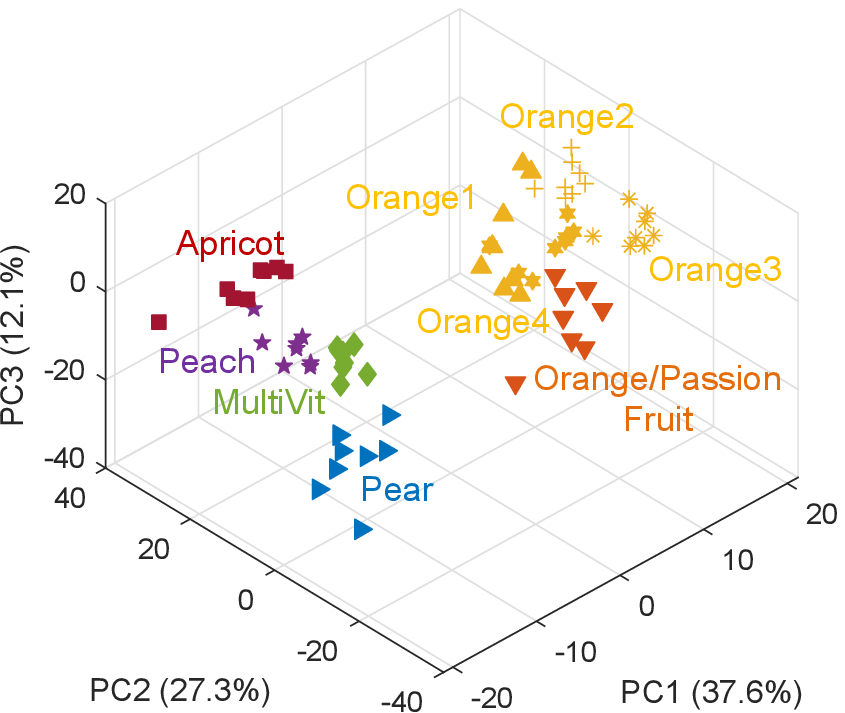}%
\label{fig_first_case}}
\hfil
\subfloat[Bottled wines]{\includegraphics[width=3in]{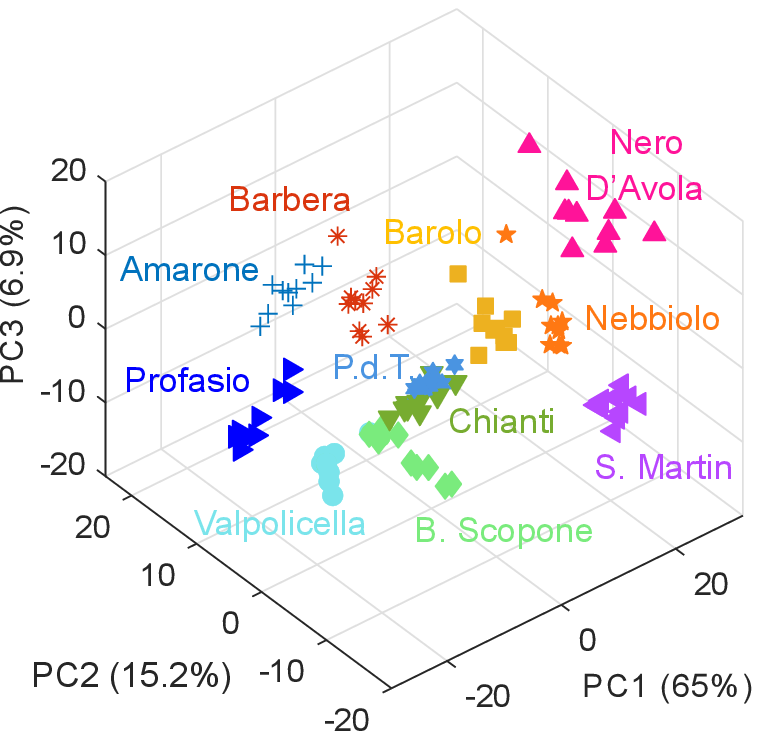}%
\label{fig_second_case}}
\caption{Principal Component Analysis (PCA) for tests with (a) 9 fruit juices and (b) and 11 bottled wines.}
\label{fig_PCAs}
\end{figure*}

Principal Component Analysis (PCA) of the 75 features extracted from transient potentiometric measurements revealed a tendency of the test data to form clusters of repeated measurements associated with the same liquid for both, fruit juices (Fig.~\ref{fig_PCAs}a) and red wines (\ref{fig_PCAs}b), respectively. Clusters of the orange juices tended to be grouped together and were more distinct from the other juices, mostly along the first principal component explaining 37.6\% of the total variance. The four sub-classes of orange juice were also resolved, despite the fact that pairs of samples (Orange1/Orange4 and Orange2/Orange3) had the same brand and nutritional values, and differed only by packaging type (Glass/PET and PET/Tetra Pak). Red wines were also clearly separated in the PC plot, with the reference wine (P.d.T) located in the middle of the plot, as expected. Remarkably, the proximity of clusters is in agreement with the region of origin: Valpolicella, Profasio, P.d.T and Amarone all originate from the Valpolicella viticulture zone in Veneto, while Barbera, Barolo and Nebbiolo are produced in Piedmont. The clusters for the two wines from Tuscany, Chianti and Brunello, are close in the PC space, while Nero d'Avola (Sicily) and Schiava (Trentino-Alto-Adige) are also separable from the others. Thus, the unsupervised analysis provides not only a clustering of the test data, but also suggests that the cross-sensitive response of the sensor array is related to intrinsic properties of the samples under test.

\subsection{Discrimination capabilities}

Classification models were trained with both data sets and compared to a baseline classifier, which was chosen to be a weighted random predictor taking into account the number of tests available for each class compared to the size of whole data set. Due to the high dimensionality of the available feature space compared to the total number of tests, PCA was used to select the minimum number of components required to explain at least 95\% of the total variance. Thus, for the classification of juices and wines, only eight and six predictors were used, respectively. Data sets were split randomly in 10 folds, whereof 9 were used for training of classification algorithms and the left-out fold was used for testing. The procedure was repeated for each of the folds and the mean accuracy across the 10 iterations is reported in the first two rows of Table \ref{Tab:Testresults} along with the corresponding standard error, computed as the ratio between the standard deviation and the square root of fold number. For both fruit juices and bottled wines, the accuracy of each of the three classification algorithms is significantly higher compared to the baseline value, indicating favorable discrimination of samples. The LDA model produced the highest classification accuracy ($>97\%$) in both cases, while the KNN algorithm was generally the least accurate ($\sim90\%$). High performance of the LDA model suggests that the underlying relationship between PCs and class separation could be easily learned by a linear model, given the reduced amount of available training instances. Furthermore, classification of wines based on their provenance was tested using the Italian region of origin as class label, namely Piedmont, Tuscany, Veneto or Other. Results based on the random 10-folds split are reported in the third row of Table \ref{Tab:Testresults}. The KNN algorithm was able to resolve the origin of wines based on the four-classes separation better (98.2$\%$) compared to the other two models (83.6$\%$ for LDA, 93.6$\%$ for bagged trees) for this task.   

\begin{table}[ht]
\centering
\caption{Classification accuracies}
\resizebox{\columnwidth}{!}{
\begin{tabular}{|c|c|c|c|c|}
\hline
{Data set} & {Baseline} & {LDA} & {KNN} & {Bag Trees}  \\  
 \hline
 Fruit Juices  & 13.9$\pm$3.9\% & \textbf{97.3$\pm$1.9\%} & 90.4$\pm$2.2\% & 94.5$\pm$3.3\% \\
\hline
 Bottled Wines & 9.1$\pm$2.0\% & \textbf{99.1$\pm$0.3\%} & 89.1$\pm$0.5\% & 90.9$\pm$0.8\%  \\
 \hline
  Wine Origin & 21.8$\pm$5.2\% & 83.6$\pm$2.5\% & \textbf{98.2$\pm$1.2\%} & 93.6$\pm$1.8\%  \\
 \hline
 Orange Flavor & 50.6$\pm$4.4\% & 88.9$\pm$9.6\% & 76.4$\pm$9.2\% & \textbf{90.3$\pm$9.7\%}  \\
 \hline
 Juice Acceptance & 51.9$\pm$7.3\% & \textbf{92.6$\pm$1.9\%} & 65.1$\pm$11.1\% & 88.4$\pm$8.4\%  \\
 \hline
 Wine Alcohol & 53.3$\pm$5.4\% & \textbf{80.9$\pm$9.7\%} & 68.2$\pm$9.5\% & 65.5$\pm$13.7\%  \\
 \hline
\end{tabular}}
\label{Tab:Testresults}
\end{table}

\subsection{Estimation of beverage attributes and sensory perception}

 A series of binary classifiers was trained to predict specific properties, such as the orange flavor of a juice, the consumer acceptability of an aged sample or the alcoholic content of a wine. For these tasks, each data set was split into a number of folds equal to the number of samples, such that each fold contained only data for the same liquid. Thus, at each iteration, it was possible to simulate the inference of an attribute of a sample \emph{unseen} during training. The prediction accuracy based on mean results across folds was always above the baseline level (rows 4--6 in Table \ref{Tab:Testresults}). High standard errors arise from misclassifications due to a particular class, corresponding to an entire mispredicted fold. LDA and the bagged tree models exhibited comparable accuracy ($\sim90\%$) for the correct prediction of the orange flavor. The main source of misclassification was the special case of the orange--passion fruit mixed juice, which was erroneously classified as an orange juice by all models. 
 Samples stored at room temperature for more than 40 days or at 40$^{\circ}$C for more than 10 days were \emph{rejected} by the sensory panel. The sensory panel response was used as ground-truth for model training on the sensor array data. The highest classification performances were obtained with the LDA model that could be used to predict the acceptance of an unknown aged juice sample with a 92.6\% accuracy. Generally, models correctly predicted the rejection of samples stored at 40$^{\circ}$C, while misclassifications occurred for samples stored for more than 40 days at room temperature. A binary classifier was also trained to predict the alcohol level of red wines, whereby each wine was assigned to a group with \textit{low} (12.5--13.5~vol\%) or \textit{high} (14--16.5~vol\%) alcohol content. This classification task was the most challenging, with the best accuracy of $80.9\%$ obtained with the LDA model. This is attributed to the relatively narrow range of alcohol vol\% for the wines under test, whereby misclassification mostly occurred for wines with intermediate alcohol content (13-13.5~vol\%).

\section{Conclusion}

A remarkably simple integrated sensor array was found to exhibit cross-sensitivity to a range of analytes, including metal ions as shown in previous work, as well as organic acids as demonstrated in the present contribution. Supervised learning was applied to various classification problems to successfully identify fruit juices by fruit type, including similar juices processed in different packages, and bottled Italian wines by their brand and origin. The electronic tongue was also successfully trained with descriptors from sensory evaluation, such as sensory panel acceptability for purchase. Due to the combination of the integrated electronic tongue with an automated data pipeline configurable via a cloud back-end or edge device such as a smartphone, its capacity to be easily reconfigured makes it attractive for remote analysis of complex liquids for a wide range of potential applications in beverage authentication, quality control and product innovation.

\section*{Acknowledgment}

\emph{The authors thank Yuksel Temiz, Keiji Matsumoto and Ralph Heller for technical contributions and Igor Bodnar, Bob Veazey, Philippe Glenat, Marco Fregonese and Alessandro Lualdi for fruitful discussions. Support from the European Commission (TrustEat 952600) is gratefully acknowledged.}

\bibliographystyle{IEEEtran}
\bibliography{HyperTaste_ISOEN.bib}

\end{document}